\documentclass[letterpaper]{article} 
\usepackage{aaai23}  
\usepackage{times}  
\usepackage{helvet}  
\usepackage{courier}  
\usepackage[hyphens]{url}  
\usepackage{graphicx} 
\usepackage{natbib}  
\usepackage{caption} 
\usepackage{multirow}
\usepackage{amsmath,amssymb}
\usepackage[ruled]{algorithm2e}
\usepackage{newfloat}
\usepackage{listings}
\DeclareCaptionStyle{ruled}{labelfont=normalfont,labelsep=colon,strut=off} 
\nocopyright 

\title{ReGANIE: Rectifying GAN Inversion Errors for Accurate Real Image Editing}
\author{
	Bingchuan Li,
	Tianxiang Ma,
	Peng Zhang,
	Miao Hua,
	Wei Liu,
	Qian He,
	Zili Yi\thanks{Corresponding author}
}

\affiliations{
	ByteDance Ltd, Beijing, China\\
	\{libingchuan, matianxiang.724, 
	liuwei.jikun, zhangpeng.ucas, heqian, huamiao, yizili\}@bytedance.com
}

\usepackage{bibentry}

\begin{document}
\maketitle

\begin{abstract}
		The StyleGAN family succeed in high-fidelity image generation and allow for flexible and plausible editing of generated images by manipulating the semantic-rich latent style space. However, projecting a real image into its latent space encounters an inherent trade-off between inversion quality and editability. Existing encoder-based or optimization-based StyleGAN inversion methods attempt to mitigate the trade-off but see limited performance. To fundamentally resolve this problem, we propose a novel two-phase framework by designating two separate networks to tackle editing and reconstruction respectively, instead of balancing the two. Specifically, in Phase I, a  $\mathcal{W}$-space-oriented StyleGAN inversion network is trained and used to perform image inversion and editing, which assures the editability but sacrifices reconstruction quality. In Phase II, a carefully designed rectifying network is utilized to rectify the inversion errors and perform ideal reconstruction. Experimental results show that our approach yields near-perfect reconstructions without sacrificing the editability, thus allowing accurate manipulation of real images. Further, we evaluate the performance of our rectifying network, and see great generalizability towards unseen manipulation types and out-of-domain images. 
	\end{abstract}
	
	\section{Introduction}
	
	As one of the flagship unconditional GANs, StyleGAN \cite{karras2019style} and its advanced versions \cite{karras2020analyzing,karras2020training,karras2021alias} are able to achieve high-fidelity image generation, but also facilitate the semantic editing in latent space. For example, some researchers  \cite{shen2020interfacegan,wu2021stylespace,patashnik2021styleclip,abdal2021styleflow,tewari2020stylerig} seek to control the attributes of generated images by manipulating the $\mathcal{W}$ or $\mathcal{S}$ latent space of StyleGAN. These methods attempt to enhance the semantic interpretability of the latent space, thus allowing the editing of the generated images. However, the real photograph needs to be embedded to StyleGAN's latent space to facilitate the editing \cite{abdal2020image2stylegan++,richardson2021encoding}, which is prone to noticeable GAN inversion errors and long-tail information loss \cite{abdal2020image2stylegan++,richardson2021encoding}.
	
	\begin{figure}[t]
		\centering
		\includegraphics[width=1\linewidth]{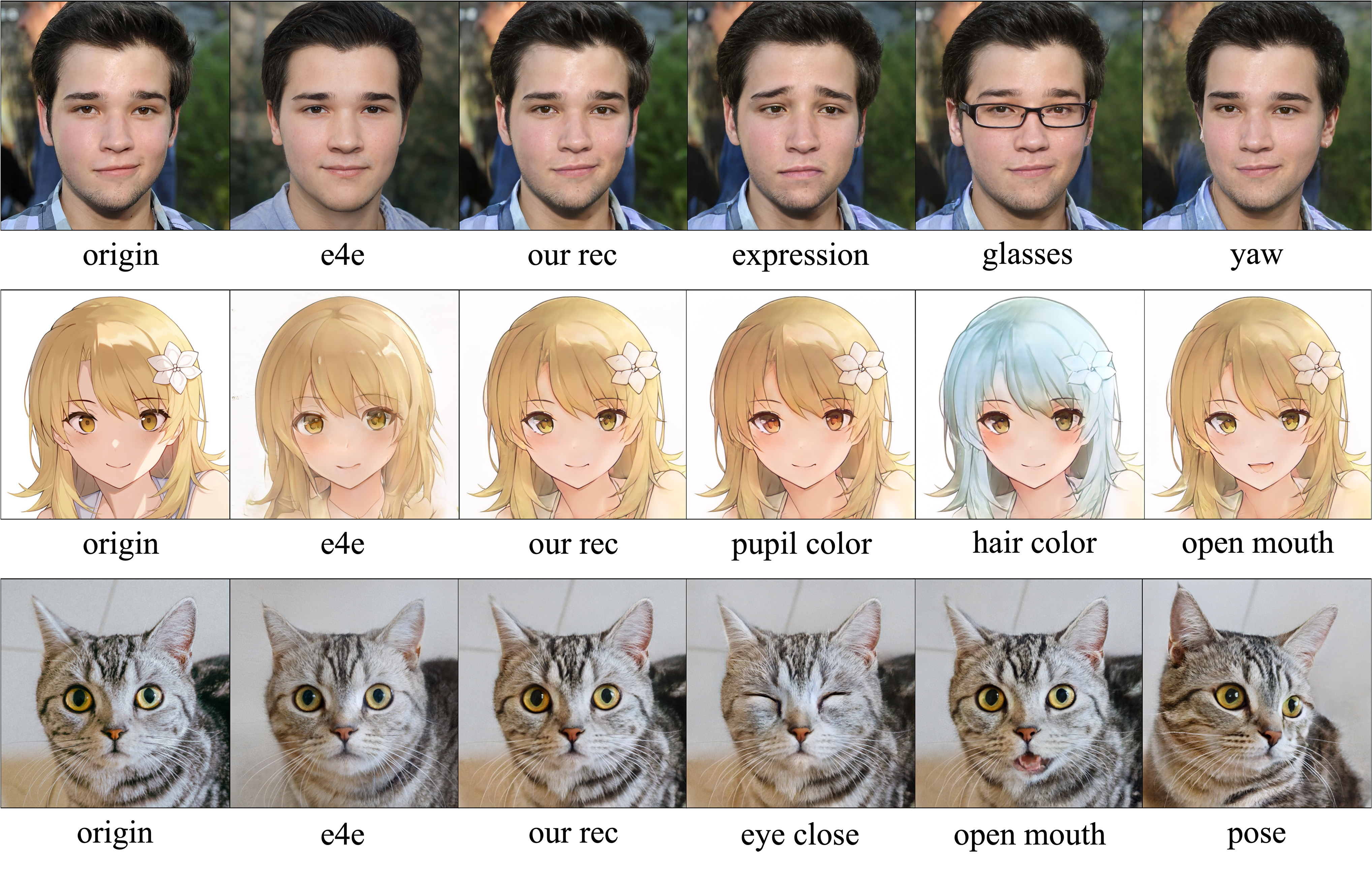}
		\captionof{figure}{The image reconstruction and attribute editing results with our approach, compared with the baseline results \cite{tov2021designing} (second column).}
		\label{fig:teaser}
	\end{figure}
	
	Recent investigations have attempted to improve reconstruction quality while maintaining the editability. These methods mainly fall into three streams. The first stream is encoder-based \cite{tov2021designing,alaluf2021restyle,dinh2022hyperinverter,alaluf2022hyperstyle,wang2022high} that optimizes the design of image encoders for improved reconstruction while keeping the StyleGAN generator fixed. The pioneering work is e4e \cite{tov2021designing}, whose encoder is carefully designed to embed an image to the $\mathcal{W}$ space. Following this work, an iterative strategy \cite{alaluf2021restyle,alaluf2022hyperstyle} to search the $\mathcal{W}$ code for better reconstruction is proposed, with the cost of increased inference time though. However, the embedding accuracy of these methods is far from being perfect. The second stream \cite{roich2022pivotal} is to fix the latent code and fine-tune the generator for better reconstruction. As the initial latent space is changed during optimization, the editability is weakened. In addition, the optimization process is time-consuming, typically two or three orders of magnitude more inference time than encoder-based methods. The third stream seeks to achieve near-perfect reconstructions by introducing spatial dimensions to the latent space \cite{kim2021exploiting}, which makes the embedding more accurate but limits the editability to ``in-place'' and semantic-implicit manipulation. That means, global editing (e.g., change the pose of an object) or attribute-conditioned editing (non-exemplar-based) of an image is intractable for these methods.

	\begin{figure*}[t]
		\includegraphics[width=.86\linewidth]{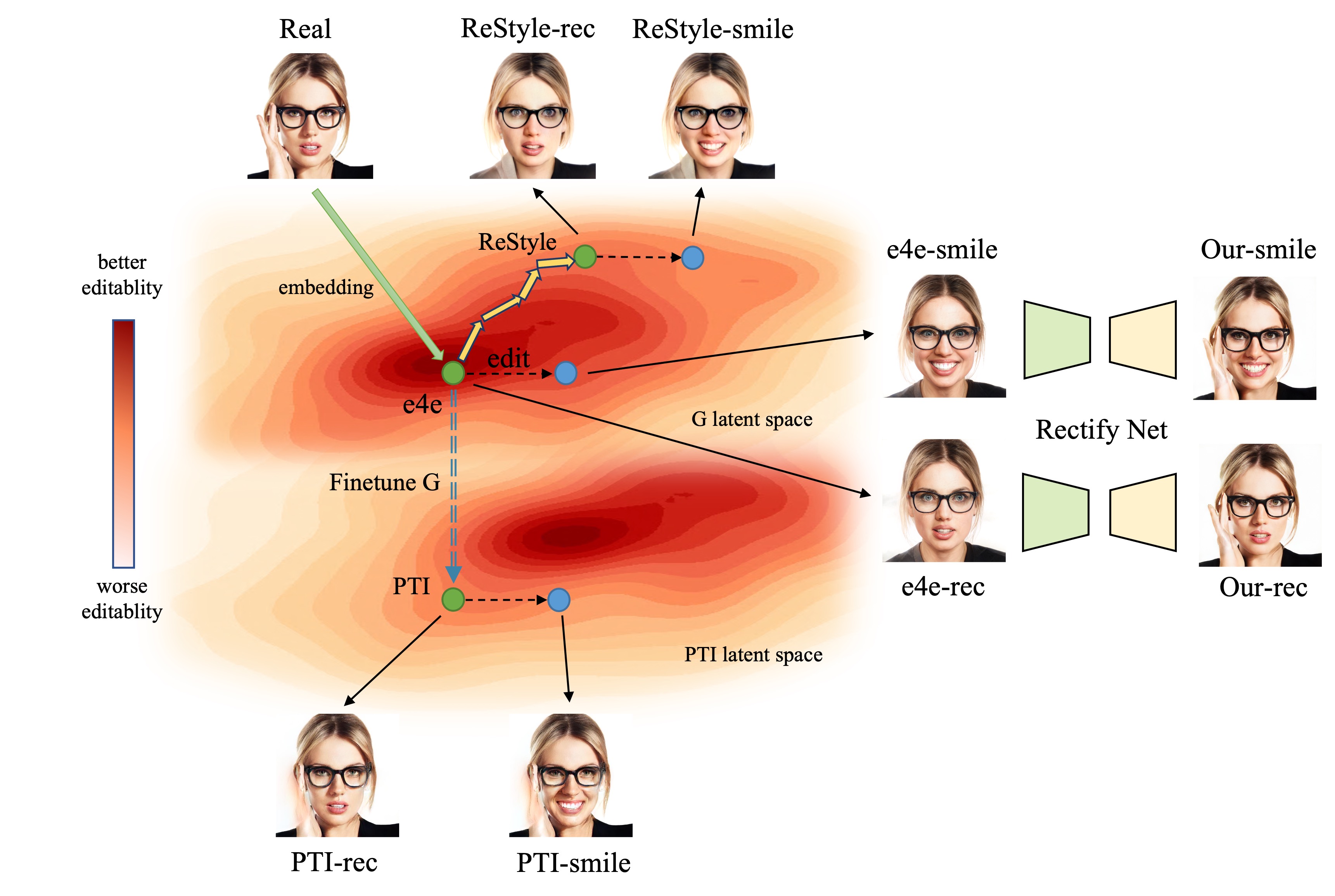}
		\centering
		\caption{Conceptual visualization of the representative StyleGAN inversion approaches and the editability-reconstruction trade-off. The heatmap highlights the editability of StyleGAN latent space ($\mathcal{W}$ or $\mathcal{S}$) \cite{karras2020analyzing,wu2021stylespace}, where darker color indicates better editability. The encoder-based approaches (e4e, ReStyle) employ either one-pass or multi-pass embedding strategy without tuning the generator. As shown, the multi-pass embedding strategy (ReStyle) produces better reconstruction but compromises editability. Optimization-based methods (PTI) seek to tune the StyleGAN generator for better reconstruction but inevitably sacrifice the editability (the latent space is shifted and distorted). Our method adopts the stable embedding and editing scheme from e4e \cite{tov2021designing} while introducing an additional rectifying network to achieve near-perfect reconstruction (hand occlusion) without	compromising the editability.}
		\label{fig:intro}
	\end{figure*}
	
	Generally, although previous schemes have made enormous efforts to improve the inversion accuracy of StyleGAN, the inherent trade-off between inversion quality and editability is not solved. Unlike previous solutions, our cleverly designed framework separates editability preservation and reconstruction improvement into two phases, thus avoiding the trade-off at all. As shown in Figure 2, we visually demonstrate the concepts and editability-reconstruction trade-offs of a representative StyleGAN inversion methods. Similar to existing StyleGAN inversion and latent-based image editing models \cite{richardson2021encoding,tov2021designing,alaluf2021restyle,alaluf2022hyperstyle}, the first phase is targeted on plausible editability and fair inversion quality. We design to generate 4 sets of paired images in the first phase for later learning to fully repair the loss of pre-inversion. The rectifying network in the second phase is designated to rectify the StyleGAN inversion errors caused in the first phase. The rectifying network takes the difference of the original image against its inversion and the initial editing result as inputs, and aims to reconstruct an ideal editing result. The difference image indicates the long-tail information loss (e.g., occlusion, accessories or other out-of-domain features) and contains sufficient supplementary information for ideal rectifying.

	Generally speaking, our contributions include:
	\begin{itemize}
		\item{For the first time, we propose REGANIE, a scheme that fundamentally addresses the editability-reconstruction trade-off for the task of latent-based real image editing.}
		\item{The unique two-phase training framework achieves near-perfect reconstruction without compromising editability, especially for the reproduction of long-tail information loss.}
		\item{We elaborate a novel architecture of the rectifying network equipped with the novel SG module, significantly leveraging the adaptability or generalizability of the rectifying network towards unseen edit types and  out-of-domain images.}
		\item{Qualitative and quantitative comparisons with extensive GAN inversion methods validate the superiority of our method in terms of the reconstruction quality and editing accuracy without significantly compromising the inference time.}
	\end{itemize}

	\section{Related Work}
	\subsection{Attribute editing}
	Attribute control of images by manipulating the latent space of StyleGAN \cite{karras2019style} is widely used. Early unsupervised methods \cite{shen2021closed,harkonen2020ganspace} apply Principal Component Analysis (PCA) on latent space or model weights, and interpretable control can be performed by layer-wise perturbation along the principal directions. Due to limited control accuracy of unsupervised methods, a number of supervised methods are investigated. For example, InterfaceGAN \cite{shen2020interfacegan} trains an SVM to discover the separation plane and editing direction for more explicit attribute control. Nonlinear methods such as  \cite{wang2021hijack,li2021dystyle,abdal2021styleflow} train neural networks to disentangle the GAN latent space by attributes. Further, the proposal of StyleSpace \cite{wu2021stylespace} and StyleCLIP \cite{radford2021learning,patashnik2021styleclip} expands the scope of semantic control. Besides, regional semantic control methods \cite{ling2021editgan,shi2022semanticstylegan,chong2021stylegan,hou2022feat} enables precise regional editing of images without unwanted global variation. Although existing StyleGAN-based image manipulation methods perform well for GAN-generated images, manipulating real images relies on accurate GAN inversion methods, which makes it challenging. Our method targets on resolving the GAN inversion errors for more accurate real image editing.
	
	\subsection{GAN inversion}
	GAN inversion is to project the image into the GAN latent space and then reconstruct it by the generator, which is widely used in image analysis, image manipulation and compression. A straightforward approach is to iteratively adjust the latent codes to minimize the difference between the generated image and the target through gradient descent. More advanced research is the encoder-based inversion methods \cite{tov2021designing,alaluf2021restyle,dinh2022hyperinverter,alaluf2022hyperstyle} that elaborate the design of image encoders for improved reconstruction accuracy. Limited reconstruction accuracy and the trade-off between reconstruction and editability are the major limitations of this stream. Some recent works \cite{wang2022high,yao2022feature,kim2021exploiting} have attempted to incorporate spatial features directly into the intermediate layers of StyleGAN generators. Although it helps improve the reconstruction, it is prone to introducing artifacts when performing the edits that involves geometric deformation or pose variation. We are motivated by the joint spatial and global features of these methods when designing our rectifying network.

	\section{Methodology}
	
	We exploit a two-phase framework to fundamentally resolve the reconstruction-editability trade-off. In the first phase, a network consisting of a StyleGAN generator, an embedding encoder and a style editor is designated to embed the input real image and perform editing in the latent space: see more details in Section Phase I. The second phase involves a rectifying network that learns to rectify the inversion error and restore missing information produced in the first phase.
	
	\begin{figure*}[t]
		\includegraphics[width=.95\linewidth]{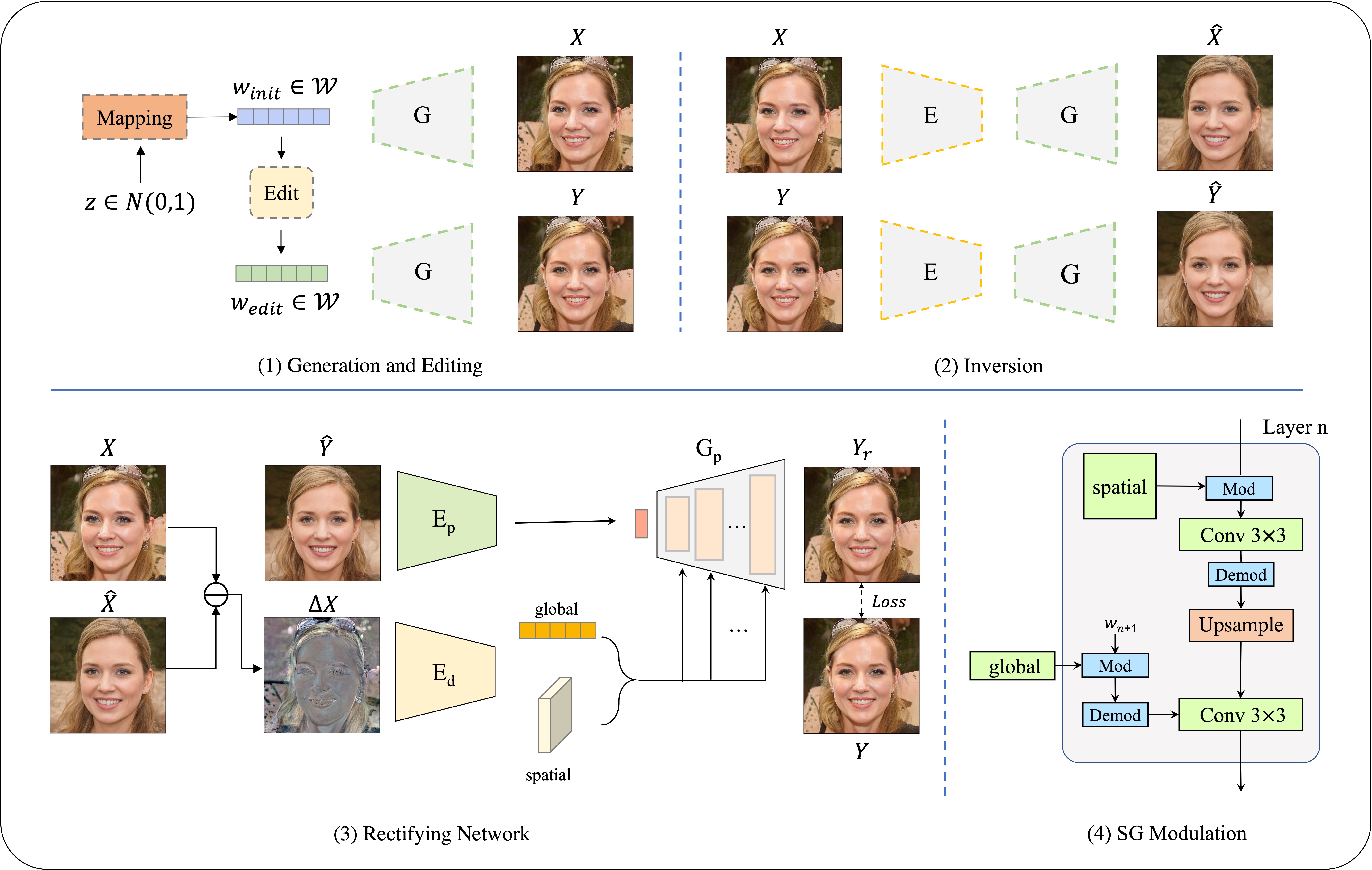}
		\centering
		\caption{The overview of our two-phase framework as for training. The upper part demonstrate the inversion and editing (e.g. pose) process of the first phase. The bottom part depicts the architecture of the rectifying network and the SG module used in Phase II. Note that the inference follows a slightly different pipeline as detailed in Alg. \ref{algorithm}.}
		\label{fig:framework}
	\end{figure*}

	\subsection{Phase I: Editing \& Pre-Inversion}
	\label{sec:editability}
	
	As shown in Figure \ref{fig:framework}, our framework relies on a pre-trained StyleGAN2. A high-fidelity image can be generated from a randomly-sampled latent vector $z_{init}$ from a normal distribution. The pre-trained mapping network $M$ maps the $z_{init}$ into $w_{init}\in \mathcal{W}$ space, and then generates an image through the generator $G$. That is 
	
	\begin{equation}
		\label{eq:i_init}
		X = G(w_{init})
	\end{equation}
	
	In general, the $\mathcal{W}$ space and its hierarchical expansion $\mathcal{W+}$ demonstrate good semantic interpretability \cite{wu2021stylespace}. $w_{init}$ can be semantically manipulated by a pre-trained style editing models \cite{shen2021closed,harkonen2020ganspace,shen2020interfacegan,patashnik2021styleclip,wang2021hijack,li2021dystyle,abdal2021styleflow,ling2021editgan,shi2022semanticstylegan,chong2021stylegan,hou2022feat}. Without losing generality, we choose the state-of-the-art non-linear multi-attribute style editor DyStyle \cite{li2021dystyle} to perform the editing. The corresponding editing result can be obtained by
	
	\begin{equation}
		\label{eq:i_edit}
		Y = G(Editor(w_{init}))
	\end{equation}
	
	
	The next task is to produce paired data for the second-stage network training, so pre-inversion of $X$, $Y$ is required. We train the encoder $E$ \cite{tov2021designing} to embed images into the latent $\mathcal{W}$ space of the original StyleGAN generator $G$, which guarantees consistency of editability in training and testing phases. Then, the generator $G$ is employed to invert the embedding of $\mathcal{W}$ space back to the image. So we can invert Image $X$ or $Y$ to its inversion as
	
	
	\begin{equation}
		\label{eq:i_inv}
		\hat{X},\ \hat{Y}  = G(E(X)),\ G(E(Y))
	\end{equation}
	
	
	\subsection{Phase II: Rectifying Network}
	\label{sec:fewshot}
	The rectifying network in Phase II targets on rectifying the GAN inversion errors and restoring the missing information produced in Phase I. The rectifying network is conditioned on the difference image between the original image $X$ and its inversion $\hat{X}$, and the inversion of the editing result $\hat{Y}$, with the expect to reconstruct the ideal editing result $Y$. 
	
	\subsubsection{Dual-pathway encoder}
	\label{sec:encoder}
	
	As shown in figure \ref{fig:framework}, the rectifying network employs a dual-pathway encoder that processes the difference image $\Delta X = X - \hat{X}$ and the primary image $\hat{Y}$ in separate branches. Since the $Y-\hat{Y}$ pair and the $X-\hat{X}$ pair are thought to suffer similar loss of information caused by the same inversion model, we assume that the difference between $X$ and $\hat{X}$ provides sufficient information for perfect reconstruction of $Y$ from $\hat{Y}$. We build separate encoders ($E_p$ and $E_d$) to process the primary image and the difference image respectively: see Figure \ref{fig:framework} (bottom). 
	
	Considering an image of resolution $512\times512$, Encoder $E_p$ and Encoder $E_d$ consist of 4 downsampling layers respectively. Specifically, $E_p$ extracts a latent feature map $f_x$ of size $32\times32$ from the primary image. $E_d$ extracts a spatial style code $sp$ of size $32\times32$ and a global style vector $gl$ of dimension $2048$, which is written as
	
	\begin{equation}
		\label{eq:sp}
		sp,\ gl = E_d(\Delta X)
	\end{equation}
	where $sp$ is the style codes with spatial dimensions, and $gl$ is the global style vector.

	\subsubsection{Generator}
	A style-based Generator $G_p$ is then used to fuse the extracted style codes ($sp$ and $gl$) and latent features ($f_x$), and generates an image with the inversion errors rectified. The generator starts from the latent feature $f_x$ and generates a $512\times512$ image by applying 4 upsampling layers. The style codes ($sp$ and $gl$) are joined to the generator through style modulation as used in \cite{park2020swapping}.  Note that $X$ and $Y$ are both generated images, they are actually exchangeable to each other, just as

	\begin{align}
		\label{eq:rec}
		X_r = G_p(E_p(\hat{X}), E_d(\Delta Y)) \\
		Y_r = G_p(E_p(\hat{Y}), E_d(\Delta X))
	\end{align}
	

	\subsubsection{Spatial and global modulation}\label{sec:spmod}
	The style codes $sp$ and $gl$ are designated to convey spatial and global information respectively. The original StyleGAN architecture projects an image from a 512-d vector that causes loss of spatial information and leads to an optimization upper bound for inversion accuracy. Therefore, we modify the generator structure to allow for joint spatial and global modulation.

	The style modulation module slightly different from that used in StyleGAN2 \cite{karras2020analyzing} to allow for alternative modulations of the spatial codes and the global vector: see Figure \ref{fig:framework} (bottomright). The equations are written as
	
	\begin{gather}
		\label{eq:spm}
		f^{l+1} = (f^{l} \cdot sp) * w^{l} + b^{l} \\
		\label{eq:glm}
		f^{l+2} = (f^{l+1} \cdot gl) * w^{l+1} + b^{l+1}
	\end{gather}
	where $f^{l}$ represents the feature map of the $l$th layer. $w^l$ and $b^l$ represent the weights and biases of the $l$th layer, respectively. ``$\cdot$'' denotes element-wise multiplication and ``$*$'' refers to convolution.
	
	Spatial intermediate features are used in  \cite{wang2022high,kim2021exploiting,yao2022feature} for improved reconstruction. However, they are prone to artifacts when  deformation occurs in the editing, as the spatial features are not well aligned with the edited image. To avoid the issue, we alternatively place the spatial modulation and global modulation in each layer of the generator. Please refer to supplement for detailed architecture. We observe that the encoder and generator will adaptively learn to re-align the spatial information based on global information.
	
	\subsubsection{Training \& inference}
	\label{sec:training}
	The rectifying network is trained with an objective consisting of three loss terms, which is defined as:
	\begin{equation}
		\label{eq:loss}
		\mathcal{L} = \lambda_{l1}\mathcal{L}_1 (I, I_r) + \lambda_{lpips}\mathcal{L}_{lpips} (I, I_r) + \lambda_{GAN}\mathcal{L}_{GAN}(I, I_r)
	\end{equation}
	where $\mathcal{L}_1$ loss is used to suppress the pixel-wise disparities between the reconstructed image $I_r$ and the original image $I$. $\mathcal{L}_{lpips}$ \cite{zhang2018unreasonable} loss measures the features similarities between the two based on a pre-trained VGG16 network \cite{simonyan2014very}, which enforces reconstructions at the feature level. The image realism is encouraged using an adversarial loss $\mathcal{L}_{GAN}$ with $R_1$ regularization \cite{karras2020analyzing}, where the discriminator $D$ is trained adversarially against the rectifying network based on Loss $\mathcal{L_D}$. 
	We set $\lambda_{GAN}=1.0$, $\lambda_{l1}=1.0$ and $\lambda_{lpips}=1.0$  in our experiments for the best performance. 
	
	\begin{algorithm}[t]
		\caption{Training \& inference}\label{algorithm}
		\textbf{Training:} \\
		\textbf{Pre-trained Models:}  $M$, $G$, $E$, $Editor$ in Phase I\\
		\textbf{Models to be optimized:}  $G_p$, $E_p$, $E_d$, $D$ in Phase II \\
		$iter = 0$; \\
		\While{iter $\leq$ N}{
			sample $z \in N(0,1)$;
			$w_{init} \leftarrow M(z)$\;
			$X = G(w_{init})$; $Y = G(Editor(w_{init}))$\;
			$\hat{X},\ \hat{Y}  = G(E(X)),\ G(E(Y))$\;
			$t \leftarrow random\_choice(\{1,\ 2,\ 3\})$\;
			\uIf{$t = 1$}{$\Delta I \leftarrow Y - \hat{Y}$; $\hat{I} \leftarrow \hat{X}$; $I \leftarrow X$;}
			\uElseIf{$t = 2$}{$\Delta I \leftarrow X - \hat{X}$; $\hat{I} \leftarrow \hat{Y}$; $I \leftarrow Y$;}
			\Else{$\Delta I \leftarrow X - \hat{X}$; $\hat{I} \leftarrow \hat{X}$; $I \leftarrow X$;}
			$I_r \leftarrow G_p(E_p(\hat{I}), E_d(\Delta I))$\;
			$\mathcal{L} \leftarrow \mathcal{L}_{rec}(I_r, I) + \mathcal{L}_{GAN}(D(I_{r}), D(I))$\;
			optimize $G_p$, $E_p$, $E_d$ based on loss $\mathcal{L}$\;
			$iter\leftarrow iter+1$\;
		}\
		
		\textbf{Inference:} \\
		$\ \ \ \ $$\hat{X},\ \hat{Y} \leftarrow G(E(X_{0})),\ G(Editor(E(X_{0})))$\;
		$\ \ \ \ $$\Delta X \leftarrow X_{0} - \hat{X}$\;
		$\ \ \ \ $$X_{r} \leftarrow G_p(E_p(\hat{X}), E_d(\Delta X))$\;
		$\ \ \ \ $$Y_{r} \leftarrow G_p(E_p(\hat{Y}), E_d(\Delta X))$\;
		\textbf{end}\\
		
	\end{algorithm}
	
	As in Algorithm \ref{algorithm}, in the first phase, we use off-the-shelf pre-trained models provided in \cite{karras2020analyzing,li2021dystyle,tov2021designing}. We train the rectifying network with generated samples from Phase I. Specifically, the training triplets ($\Delta I$, $\hat{I}$ and $I$) can be obtained in three ways to encourage reconstructions of both edit-free or edit-involved scenarios. First, $\Delta I = Y-\hat{Y}$, $\hat{I}=\hat{X}$ and $I=X$. Second, $\Delta I = X-\hat{X}$, $\hat{I}=\hat{Y}$ and $I=Y$. Third, $\Delta I = X-\hat{X}$, $\hat{I}=\hat{X}$ and $I=X$. The former two encourage reconstructing from its inversion and the inversion errors of its edited partner. The third way encourages reconstructing from its own inversion and inversion errors. During training, the attribute set intended for manipulation are those provided by DyStyle \cite{li2021dystyle} network, including pose, age, glasses, expression, etc.
	
	During inference, the rectifying network is expected to reconstruct an ideal edited image $Y_r$ from its inversion $\hat{I}=\hat{Y}$ and the inversion errors of the input image $\Delta I = X_o -\hat{X}$. It is also encouraged to reconstruct the input image itself $X_r$ from its own inversion $\hat{I}=\hat{X}$ and inversion errors $\Delta I = X_o -\hat{X}$. 
	
	\section{Experiments}
	
	\subsection{Experimental setup}
	The pre-trained StyleGAN models that our method relies on are typically trained on facial datasets. To demonstrate the generalizability of our method on unseen realistic faces, we train our model on the FFHQ \cite{karras2019style} dataset, and test on the CelebA-HQ \cite{karras2018progressive} dataset. We choose the Dystyle \cite{li2021dystyle} and StyleCLIP as the latent editor, as the pre-trained models for 
	multi-attribute-conditioned or text-conditioned editing are available. We also experiment our method on the animal portrait dataset \cite{choi2020stargan} and the anime dataset \cite{branwen2019danbooru2019}.
	
	Without losing generality, we adopt the GAN inversion model and training procedure of the embedding encoder from e4e \cite{tov2021designing}. We re-trained the embedding encoder for the anime and animal portrait datasets. In addition, we also compare our method with PTI \cite{roich2022pivotal}, a representative optimization-based method. The official pre-trained models provided in PTI is used as the baseline for evaluation.
	
	\subsection{Implementation details}
	The proposed ReGANIE network is trained on a single Tesla V100 GPU with batch size of 8 on $512\times512$ resolution images. It is optimized using the Adam optimizer \cite{kingma2014adam} with b1 = 0.0 and b2 = 0.99, and the learning rate is fixed at 0.002. In all experiments, the training is performed for 100,000 iterations. The inference costs 190ms on a single $512\times512$ image.
	
	\subsection{Evaluations of image reconstruction quality}
	\begin{figure}[t]
		\includegraphics[width=0.98\columnwidth]{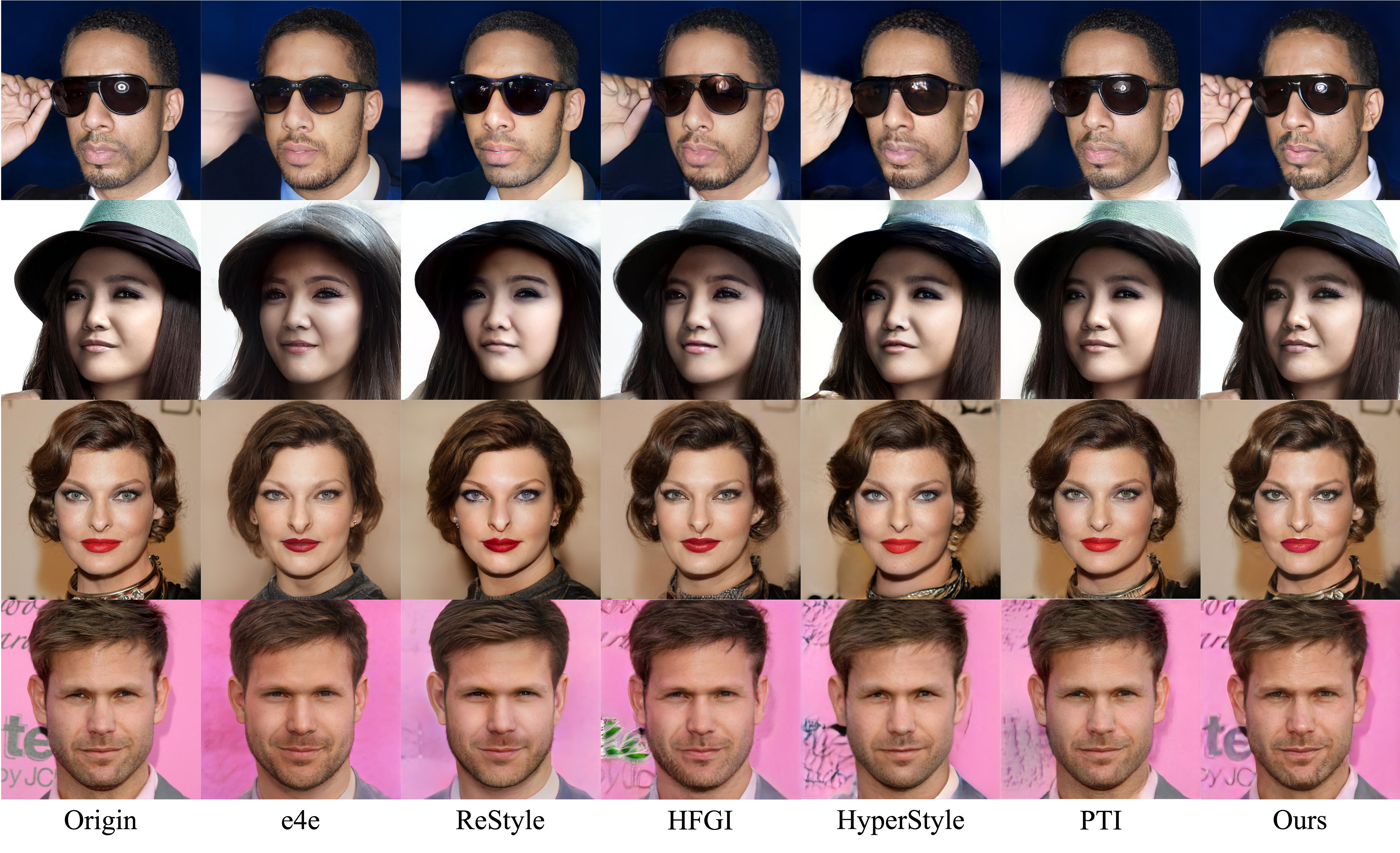}
		\centering
		\caption{Comparison of reconstruction quality. Four different long-tail loss subjects are demonstrated, including hand and glasses occlusion (1st row), hats (2nd row), accessories and makeup (3rd row), and background (4th row). }
		\label{fig:compare_rec}
	\end{figure}
	\subsubsection{Qualitative results}
	To verify the reconstruction accuracy of the proposed method, we compare our approach with state-of-the-art GAN inversion methods including e4e \cite{tov2021designing}, ReStyle \cite{alaluf2021restyle}, HFGI \cite{wang2022high}, HyperStyle \cite{alaluf2022hyperstyle} and PTI \cite{roich2022pivotal}: see Figure \ref{fig:compare_rec}. 
	
	We can observe that the encoder-based methods e4e and ReStyle cannot restore difficult cases well (long-tail information loss). HFGI, HyperStyle and PTI are able to recover more details in the original image due to use of spatial information encoding or generator optimization, but the inversion accuracy is far from being perfect. Our approach achieves the best reconstruction accuracy and the reconstruction errors are mostly unnoticeable. We provide user study and more visualizations, please see the supplementary material.
	
	\subsubsection{Quantitative results}
	\setlength{\tabcolsep}{4pt}
	\begin{table}[t]
		\begin{center}
			\caption{Quantitative comparison of reconstruction quality on CelebA-HQ \cite{karras2018progressive}.}
			\label{table:compare_rec}
			\resizebox{\columnwidth}{!}{ 
				\begin{tabular}{l|cccc|c}
					\hline\noalign{\smallskip}
					Methods & L2($\downarrow$) & LPIPS($\downarrow$) & ID($\uparrow$) &MS-SSIM($\uparrow$) & Times($\downarrow$)\\
					\noalign{\smallskip}
					\hline
					\noalign{\smallskip}
					e4e & 0.048 & 0.2 & 0.61 & 0.73 & \bf0.05\\
					ReStyle & 0.043 & 0.18 & 0.63 & 0.79 & 0.46\\
					HFGI & 0.039 & 0.16 & 0.78 & 0.83 & 0.24\\
					HyperStyle & 0.021& 0.11 & 0.83 & 0.85 & 1.13\\
					PTI & 0.019 & 0.08 & 0.84 & 0.9 & 76\\
					\hline
					\noalign{\smallskip}
					Ours & \bf 0.016 & \bf 0.07 & \bf 0.85 & \bf 0.9 & 0.19\\
					\hline
				\end{tabular}
			}
		\end{center}
	\end{table}
	\setlength{\tabcolsep}{1.4pt}
	We perform a quantitative comparison of related GAN inversion methods and ours in terms of the reconstruction accuracy and inference time: see Table \ref{table:compare_rec}. The similarity between the reconstructed image and the original was measured with L2 distance, LPIPS \cite{zhang2018unreasonable} and MS-SSIM \cite{wang2003multiscale} scores. Additionally, the identity similarity is measured using a pre-trained face recognition model provided by \cite{huang2020curricularface}. Note that, unlike other methods, we do not explicitly apply the identity preserving loss during training. Finally, the inference time is also tested. e4e \cite{tov2021designing}, HFGI \cite{wang2022high} and our method are single-pass methods, while ReStyle \cite{alaluf2021restyle} and HyperStyle \cite{alaluf2022hyperstyle} employ multi-pass forwarding (5 based on the official recommendation). Since PTI \cite{roich2022pivotal} is optimization-based, it is iterated for 450 times to search the initial code, followed by 350 iterations to update the generator weights. PTI costs 2-3 orders of magnitude more time than the encoder-based methods. As shown in Table~\ref{table:compare_rec}, our method  outperforms other methods in terms of the reconstruction accuracy by large margin, while the speed is among the best two.
	
	\subsection{Evaluations of real image editing}
	\subsubsection{Qualitative results}
	\begin{figure}[t]
		\includegraphics[width=\columnwidth]{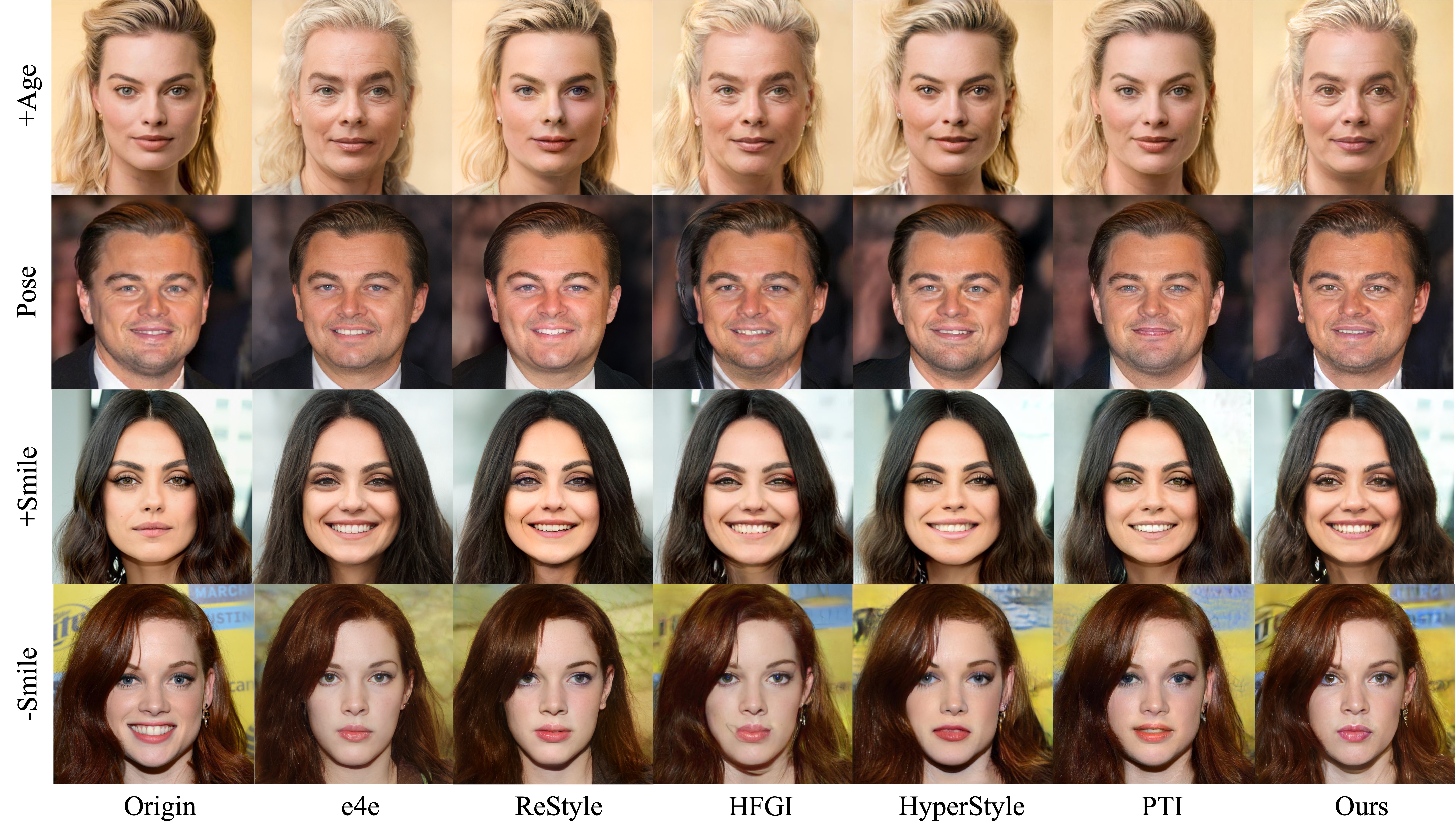}
		\centering
		\caption{Comparisons of the real face editing results while manipulating the age, pose and expression.}
		\label{fig:compare_edit}
	\end{figure}
	To visually compare different image editing methods, we demonstrate their results as for manipulating three types of facial attributes: age, pose and expression. The selected attributes are representatives of three typical editing scenarios. Specifically, age involves whole-image texture editing, pose involves global geometric deformation, and expression involves local deformation. As our rectifying network is trained with the editing results of DyStyle \cite{li2021dystyle}, for fair comparisons, we use InterfaceGAN \cite{shen2020interfacegan} as the style editor in the test phase for all methods.

	It can be observed from Figure \ref{fig:compare_edit} that ReStyle \cite{alaluf2021restyle}, HyperStyle \cite{alaluf2022hyperstyle}, and PTI \cite{roich2022pivotal} see limited attribute control accuracy: see the failure to produce wrinkles while designated for aging. HFGI \cite{wang2022high} produces silhouette artifacts, which can be caused by its naive injection of spatial information into the generator. The e4e \cite{tov2021designing} model obtains stable editing results, but suffer noticeable reconstruction errors and unwanted information loss. In general, our method achieves the best reconstruction quality while not compromising the attribute control accuracy. User study and more visualizations see the supplement.
	
	\subsubsection{Qualitative results}
	\setlength{\tabcolsep}{4pt}
	\begin{table}[t]
		\begin{center}
			\caption{Quantitative comparisons of attribute editing accuracy on CelebA-HQ \cite{karras2018progressive}. The attribute control respondence is indicated by the perceived attribute variation given the fixed editing magnitude $\alpha$ (the greater absolute value the better).}
			\label{table:compare_edit}
			\resizebox{\columnwidth}{!}{ 
				\begin{tabular}{l|c|rrrrrr}
					\hline\noalign{\smallskip}
					Attr. & $\alpha$ & e4e & ReStyle & HFGI & HyperStyle & PTI & Ours\\
					\noalign{\smallskip}
					\hline
					\noalign{\smallskip}
					\multirow{3}*{Age}& $\alpha$=-2 & -11.53 & -8.37 & -11.64 & -9.73 &-9.05 &\bf-12.18\\
					~ & $\alpha$=2 & 9.02 & 6.42 & 9.25 & 8.46& 8.11&\bf10.59\\
					\cline{2-8}\noalign{\smallskip}
					~ & Id & 0.44 & 0.45 & 0.48 & 0.59& 0.57 & \bf0.62\\
					\hline\noalign{\smallskip}
					\multirow{3}*{Yaw}& $\alpha$=-2 & -9.12& -6.65 & -7.56 & -8.9 & -8.02&\bf-9.13\\
					~ & $\alpha$=2 & 9.52& 6.75 & 7.83 & 9.06 & 8.37&\bf9.54\\
					\cline{2-8}\noalign{\smallskip}
					~ & Id & 0.52 & 0.54 & 0.61 & 0.72& 0.69 & \bf0.75\\
					\hline
				\end{tabular}
			}
		\end{center}
	\end{table}
	\setlength{\tabcolsep}{1.4pt}
	We quantitatively evaluate the editing quality of different methods, by measuring the attribute control accuracy and identity preservation. As InterfaceGAN edits image attributes by linearly manipulating the StyleGAN latent space $\mathcal{W}$. Specifically, $w_{edit} = w_{init}  + \alpha * d$, where ${d}$ is the normal vector of the separation plane (or edit direction) for a specific attribute,  and $\alpha$ is the edit magnitude. The attribute of an image can be continuously manipulated by controlling the value of $\alpha$. 
	
	We use two numeric attributes, age and yaw angle, for quantitative evaluation. We intend to see how the perceived attribute value responds to the value of $\alpha$. It is believed that the greater the perceived attribute value varies given the same amount of $\alpha$ shift, the better the editability is. We employ the official pre-trained HopeNet \cite{ruiz2018fine} model for pose estimation, and the age regressor is trained by ourselves based on the official implementation of Dex VGG \cite{alaluf2021only}. In addition, we also calculated the average identity similarity \cite{huang2020curricularface} score as the indicator of identity preservation. As shown in Table \ref{table:compare_edit}, our method and e4e \cite{tov2021designing} achieve the best attribute control competency, while our method achieve the best identity preservation, implying that our method faithfully manipulates the target attribute without producing unwanted changes along other attributes.

	\subsection{Ablation study}
	\label{sect:ablation}
	\begin{figure}[t]
		\includegraphics[width=\columnwidth]{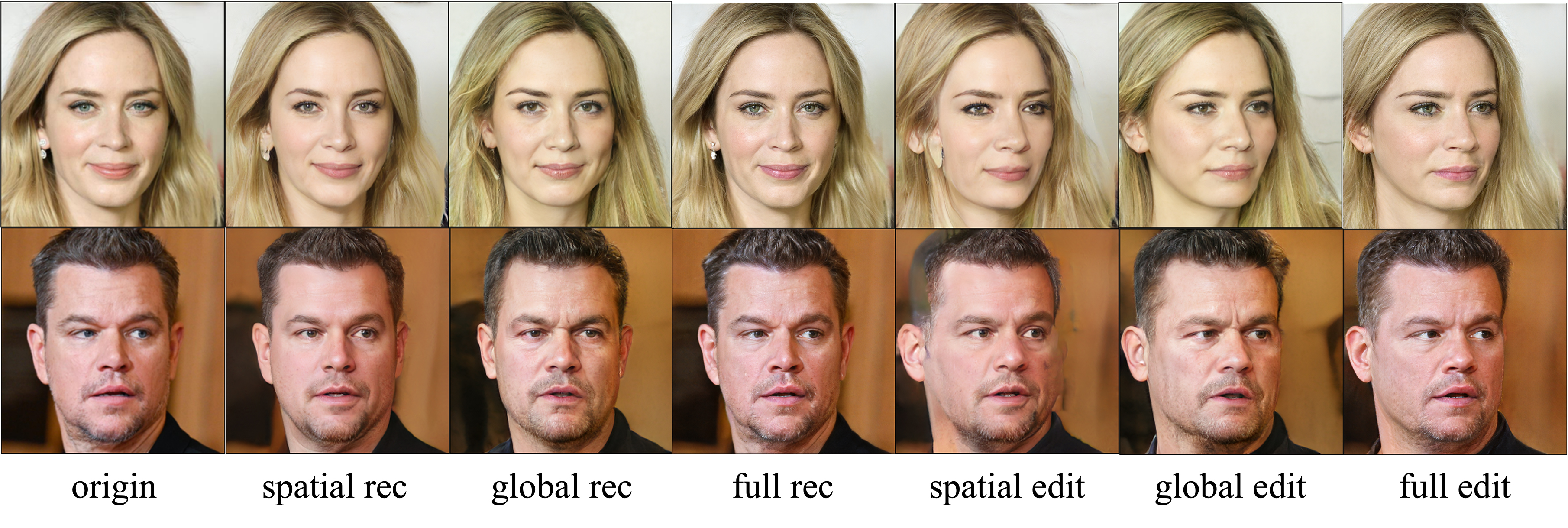}
		\centering
		\caption{Ablation studies on the spatial and global modulation (SG) module for image reconstruction and pose editing.}
		\label{fig:spmod}
	\end{figure}
	We compare the effects of spatial and global modulations mentioned on image reconstruction and editing. As shown in Figure \ref{fig:spmod}, using the global modulation only (3rd column) sees missing details such as earrings and hairstyles. In contrast, using spatial modulation only improves the reconstruction quality, but produces noticeable artifacts for attribute editing where deformation happens (5th column). The rational design of the SG module that combines the spatial and global modulation performs the best in reconstruction (4th column) and editing (7th column). Furthermore, we conduct additional ablation studies on the input of the rectifying network. Please refer to supplementary materials for detailed experimental analysis.

	\subsection{Generalizability}
	\label{sect:applications}
	\subsubsection{Unseen manipulations}
	\begin{figure}[t]
		\includegraphics[width=\columnwidth]{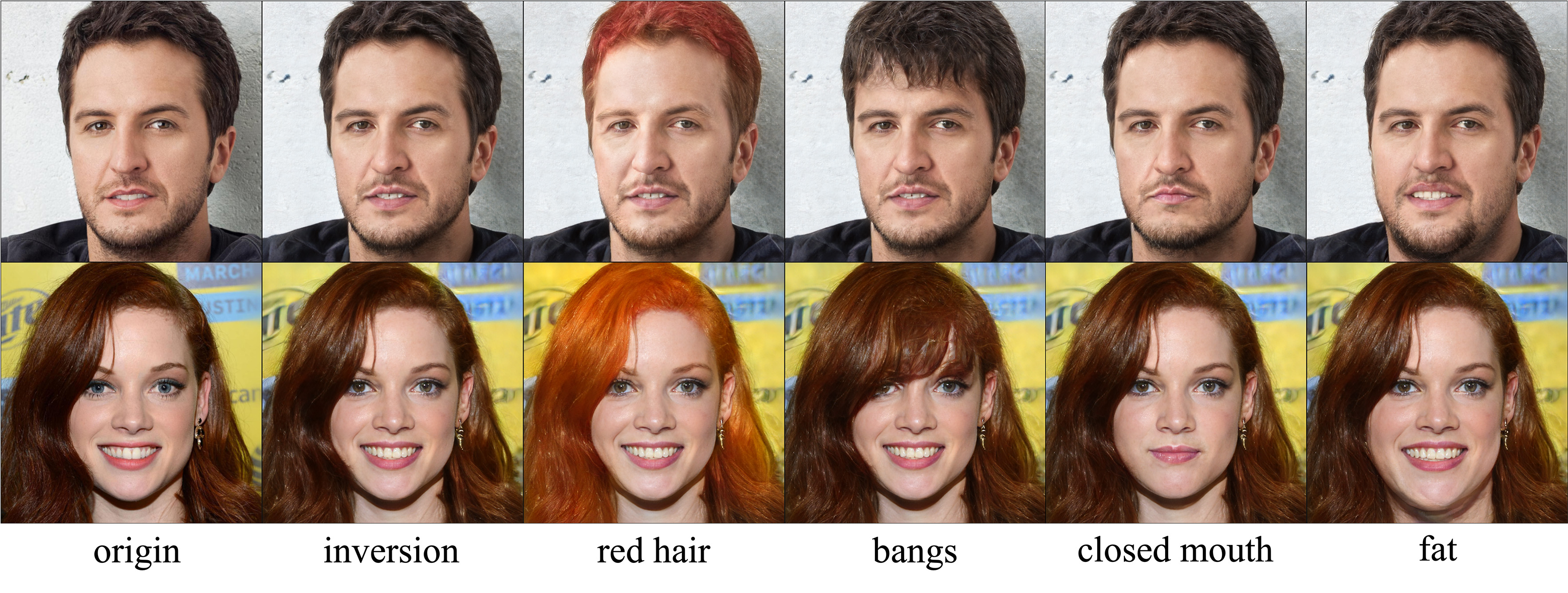}
		\centering
		\caption{The performance of our rectifying network as used to reconstruct the results of unseen attribute manipulations by StyleCLIP \cite{patashnik2021styleclip}.}
		\label{fig:fewshot}
	\end{figure}

	In this work, we use DyStyle \cite{li2021dystyle} to generate the editing results required for the training of the rectifying network. However, what is inspiring is that the performance of the rectifying network for manipulations of unseen attributes such as hair color, bangs, or face shape is also plausible: see Figure \ref{fig:fewshot}. This implies that the generator has learned to align the two inputs regardless of the edit types.
	
	\subsubsection{Out-of-domain images}
	\begin{figure}[t]
		\includegraphics[width=\columnwidth]{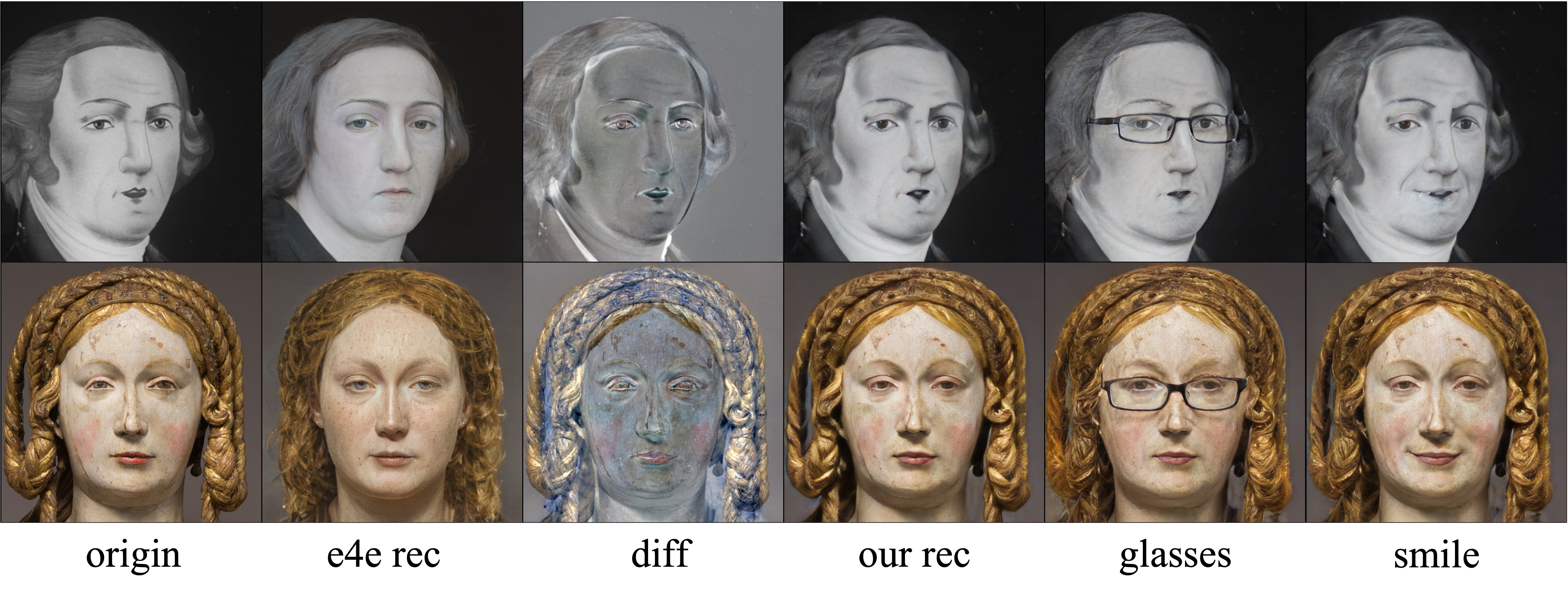}
		\centering
		\caption{The GAN inversion error rectifying results on the Metface \cite{karras2020training} dataset.}
		\label{fig:metface}
	\end{figure}
	All networks in Phase I and Phase II are trained with images from the same domain. However, we discover that our rectifying network works well on certain unseen images out of the target domain. As shown in Figure \ref{fig:metface}, the editing results on the artistic portrait dataset can be rectified with the rectifying network trained on FFHQ dataset without any fine-tuning. This implies that our rectifying network has learned to fuse and repair the inputs without overfitting into a specific domain.
	
\section{Conclusions}
We propose ReGANIE, a novel two-phase framework for accurate latent-based realistic image editing. Compared to previous encoder-based or optimization-based inversion-and-editing methods that perform reconstruction and editing with one generator, we successfully resolve the reconstruction-editability trade-off by designating separate networks to deal with the editing and reconstruction respectively. As a result, we achieve the most accurate real image editing results, without significantly increasing the inference time. Furthermore, ReGANIE exhibits great generalization towards unseen manipulation types (e.g., unseen attributes of editing and certain out-of-domain images). Finally, our framework is scalable and can be customized by choosing different GAN inversion methods or style editors for specific scenarios.

\clearpage
\bibliography{aaai23}

\clearpage
\section*{Appendix}
\subsection*{A. Architecture of the Rectifying Network }
We describe the difference between the SG modulation and the original StyleGAN2 \cite{karras2020analyzing} modulation in Sec. \ref{sec:spmod}, which is briefly shown in Fig. \ref{fig:framework}. Fig. \ref{fig:sup_sgmod} shows the detailed SG modulation block, which is a residual structure composed of spatial modulation and two sequential global modulations. We also introduce the Rectifying Network architecture in Table \ref{table:sup_arch}, taking an input image of $512\times512$ resolution. The encoder $E_p$ and $E_d$ share the backbone, where $E_d$ has one additional head for outputting the global style. The generator part mainly consists of stacked SG modulation blocks (Fig. \ref{fig:sup_sgmod}) and the last Global modulation is replaced with the ToRGB operation \cite{karras2019style}. The discriminator architecture is the same as StyleGAN2 \cite{karras2020analyzing}.

\subsection*{B. User study}
\begin{figure}[h]
	\includegraphics[width=\columnwidth]{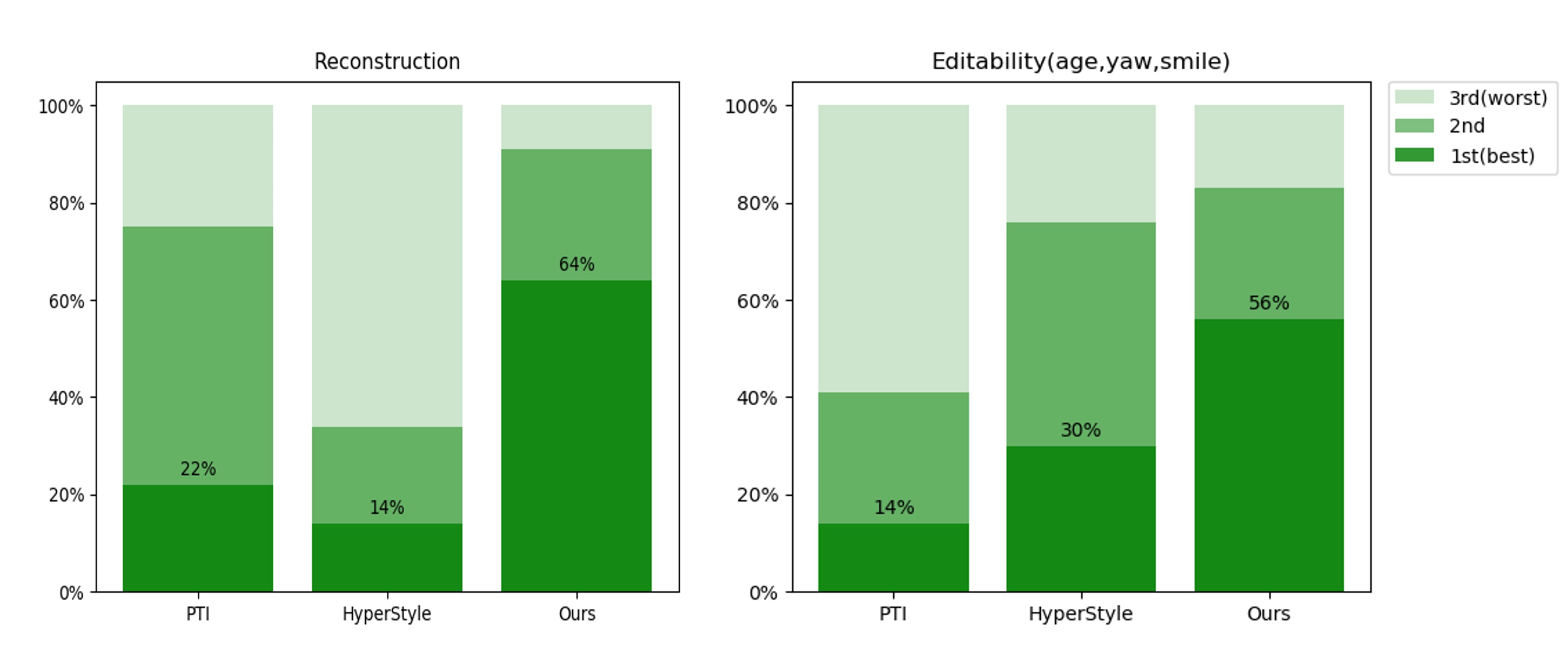}
	\centering
	\caption{User study results with SOTA optimization-based method (PTI \cite{wang2022high}) and encoder-based method (HyperStyle \cite{alaluf2022hyperstyle}).}
	\label{fig:user_study}
\end{figure}

To perceptually evaluate reconstruction and editing performance, we performed a user study.  PTI  \cite{wang2022high} and HyperStyle \cite{alaluf2022hyperstyle} were chosen for comparison as SOAT optimization-based inversion method and encoder-based inversion method, respectively. We collected 1000 votes from 20 participants evaluating 50 images, with each vote recording the method from best to worst performing. As shown in the statistical result (Figure \ref{fig:user_study}), our method gets the first approval of most people, whether it is reconstruction or editing performance.

\subsection*{C. Extended ablations studies}
\begin{figure}[h]
	\includegraphics[width=\columnwidth]{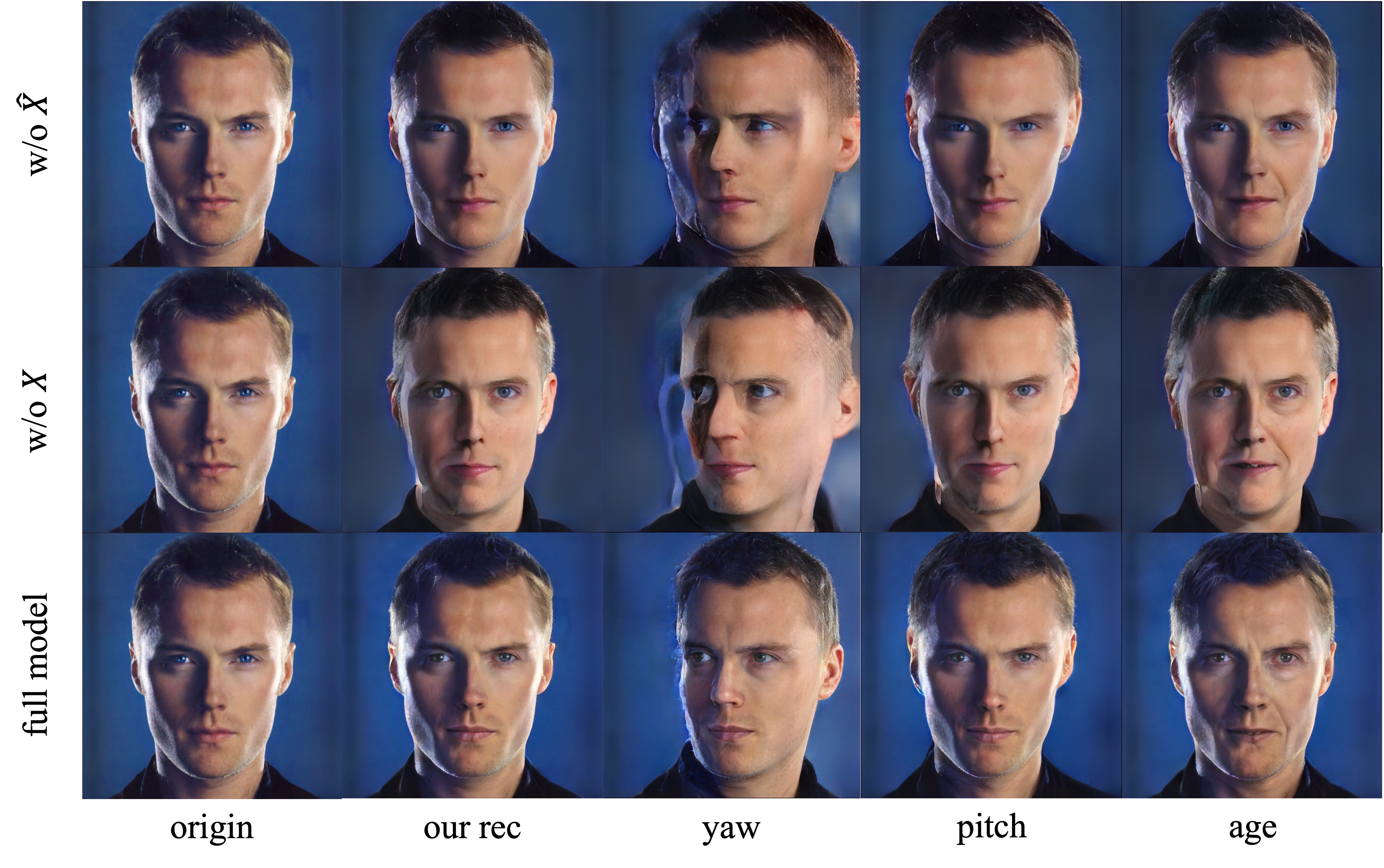}
	\centering
	\caption{Ablation studies for rectifying network inputs.}
	\label{fig:ab2}
\end{figure}
We propose a novel two-stage inversion framework, where the rectifying network is provided with input information of three images and one image for supervised training. We perform an ablation study on the input of the rectifying network, replacing the residual input $\Delta X$ with $X$ and $\hat{X}$, respectively. As shown in Figure \ref{fig:ab2}, only taking $X$ as the conditional input will cause deformation artifacts, especially the change of pose.  Only taking $\hat{X}$ as the conditional input will lose the original image information as a guide, which greatly deteriorates the reconstruction similarity. Clearly, the complete input gets stable performance. This also shows that the pre-inversion design that generates 4 sets of images is crucial.

\subsection*{D. Visual experimental results}
More qualitative comparisons on the human face dataset are shown in Fig. \ref{fig:sup_face} and Fig. \ref{fig:sup_video}, where video inversion reflects the temporal consistency of details across frames. Fig. \ref{fig:sup_cat} and Fig. \ref{fig:sup_anime} show the reconstruction and editing of animal faces and anime faces, respectively. We also validate out-of-domain image performance on a multi-style portrait dataset in Fig. \ref{fig:sup_domain} without any fine-tuning. The proposed method not only reconstructs near-perfect performance without compromising editability.

\subsection*{E. Limitation analysis}
Experiments on the anime dataset \cite{branwen2019danbooru2019}  demonstrate the limitations of our method in reconstruction as shown in Figure \ref{fig:sup_badcase}. We utilize StyleGAN2 \cite{karras2020analyzing} to randomly generate some fake images (truncation=1) and reconstruct them like real images. The results show that the fake image can be reconstructed almost perfectly despite the serious homogeneity, while the real image has a visual reconstruction bias. Comparing the same almost perfectly reconstructed human face \cite{karras2019style} and cat face \cite{choi2020stargan}, the FID \cite{heusel2017gans} indicators of StyleGAN2 converge to 3.8 and 5.5 respectively, while the Fréchet Inception Distance (FID) of the anime dataset is 11.9. A larger FID indicates a larger gap between the real image distribution and the generated image distribution, which makes our method prone to overfitting to fake image styles. Due to the high diversity of the anime dataset and the lack of strict alignment criteria, it is difficult for StyleGAN to cover all its styles. Future unconditional generators with more stylistic diversity would probably relieve this limitation.

\clearpage 	
\begin{figure*}[t]
	\includegraphics[width=0.9\linewidth]{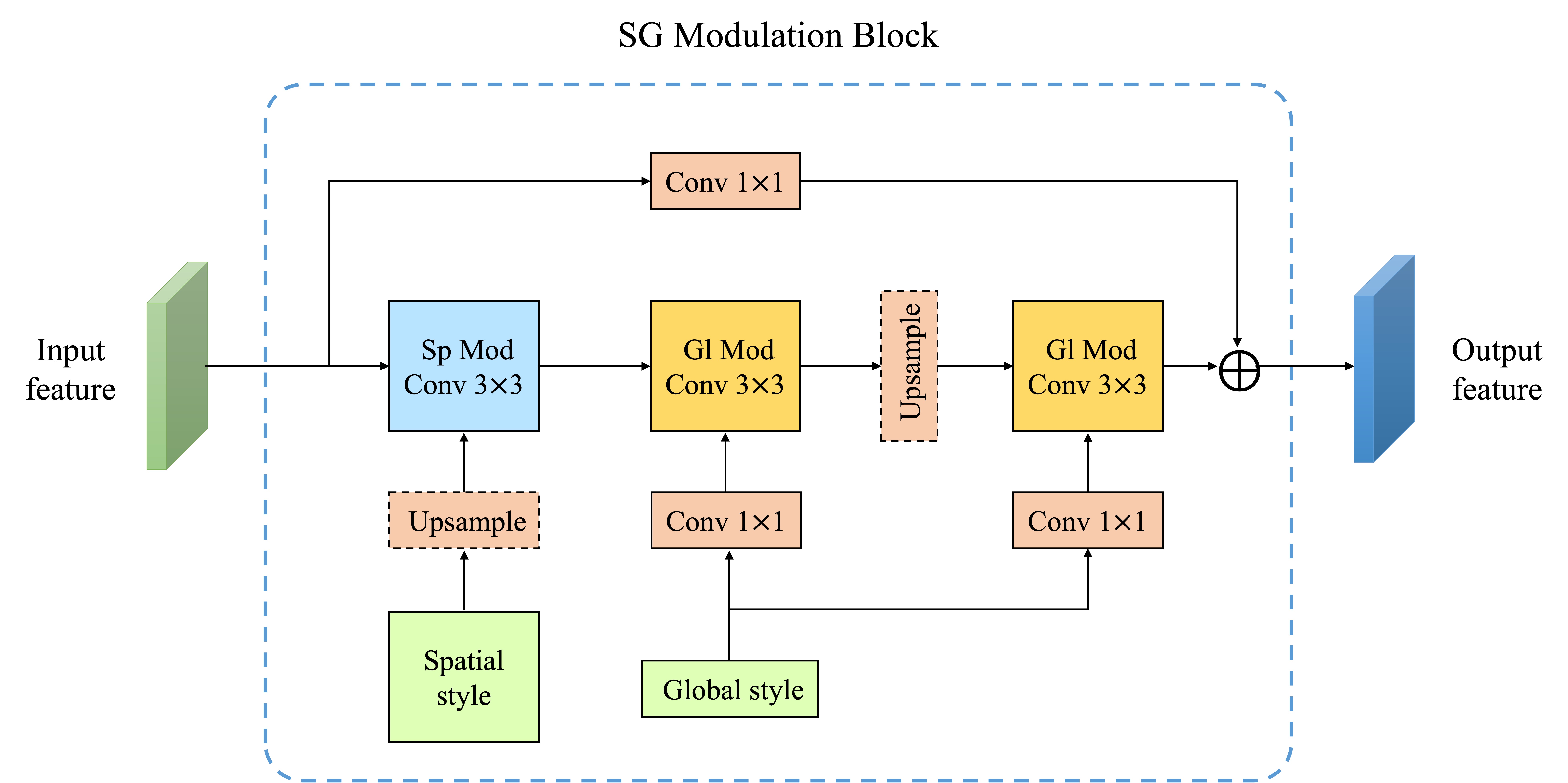}
	\centering
	\caption{Details of the SG modulation block, where the up-sampling operation resizes the input feature to target size as needed.}
	\label{fig:sup_sgmod}
\end{figure*}
\setlength{\tabcolsep}{15pt}
\begin{table*}[b]
	\begin{center}
		\caption{The detailed architecture of the Rectifying Network. $\uparrow$ (or $\downarrow$) indicate the up-sampling (or down-sampling) operation inside a layer or a block.}
		\label{table:sup_arch}
		\resizebox{0.9\linewidth}{!}{ 
		\begin{tabular}{c|c|c}
			\hline\noalign{\smallskip}
			Module & Layers & Output size\\
			\noalign{\smallskip}
			\hline
			\noalign{\smallskip}
			\multirow{5}*{Encoder (Backbone)}& Conv $3\times3$ & $512\times512\times32$ \\
			~ & ResidualBlock ($\downarrow$)  \cite{he2016deep}& $256\times256\times64$ \\
			~ & ResidualBlock ($\downarrow$) & $128\times128\times128$ \\
			~ & ResidualBlock ($\downarrow$) & $64\times64\times256$ \\
			~ & ResidualBlock ($\downarrow$) & $32\times32\times512$ \\
			\hline\noalign{\smallskip}
			\multirow{2}*{$E_p$ or $E_d$ (Spatial head)}& Conv $3\times3$ & $32\times32\times512$ \\
			~ & Conv $1\times1$ & $32\times32\times16$\\
			\hline\noalign{\smallskip}
			\multirow{3}*{$E_d$ (Global head)}& Conv $3\times3$ ($\downarrow$) & $15\times15\times1024$ \\
			~ & Conv $3\times3$ ($\downarrow$) & $7\times7\times2048$ \\
			~ & Conv $1\times1$ & $1\times2048$ \\
			\hline\noalign{\smallskip}
			\multirow{8}*{Generator}& Gl Mod Conv $3\times3$ & $32\times32\times16$ \\
			~ & SG Mod Conv $3\times3$  & $32\times32\times256$ \\
			~ & SG Mod Conv $3\times3$  & $32\times32\times512$ \\
			~ & SG Mod Conv $3\times3$ ($\uparrow$) & $64\times64\times512$ \\
			~ & SG Mod Conv $3\times3$ ($\uparrow$) & $128\times128\times512$ \\
			~ & SG Mod Conv $3\times3$ ($\uparrow$) & $256\times256\times256$ \\
			~ & SG Mod Conv $3\times3$ ($\uparrow$) & $512\times512\times128$ \\
			~ & Gl Mod Conv $1\times1$  & $512\times512\times3$ \\
			\hline
		\end{tabular}}
	\end{center}
\end{table*}
\setlength{\tabcolsep}{1.4pt}

\clearpage

\begin{figure*}[p]
	\includegraphics[width=\linewidth]{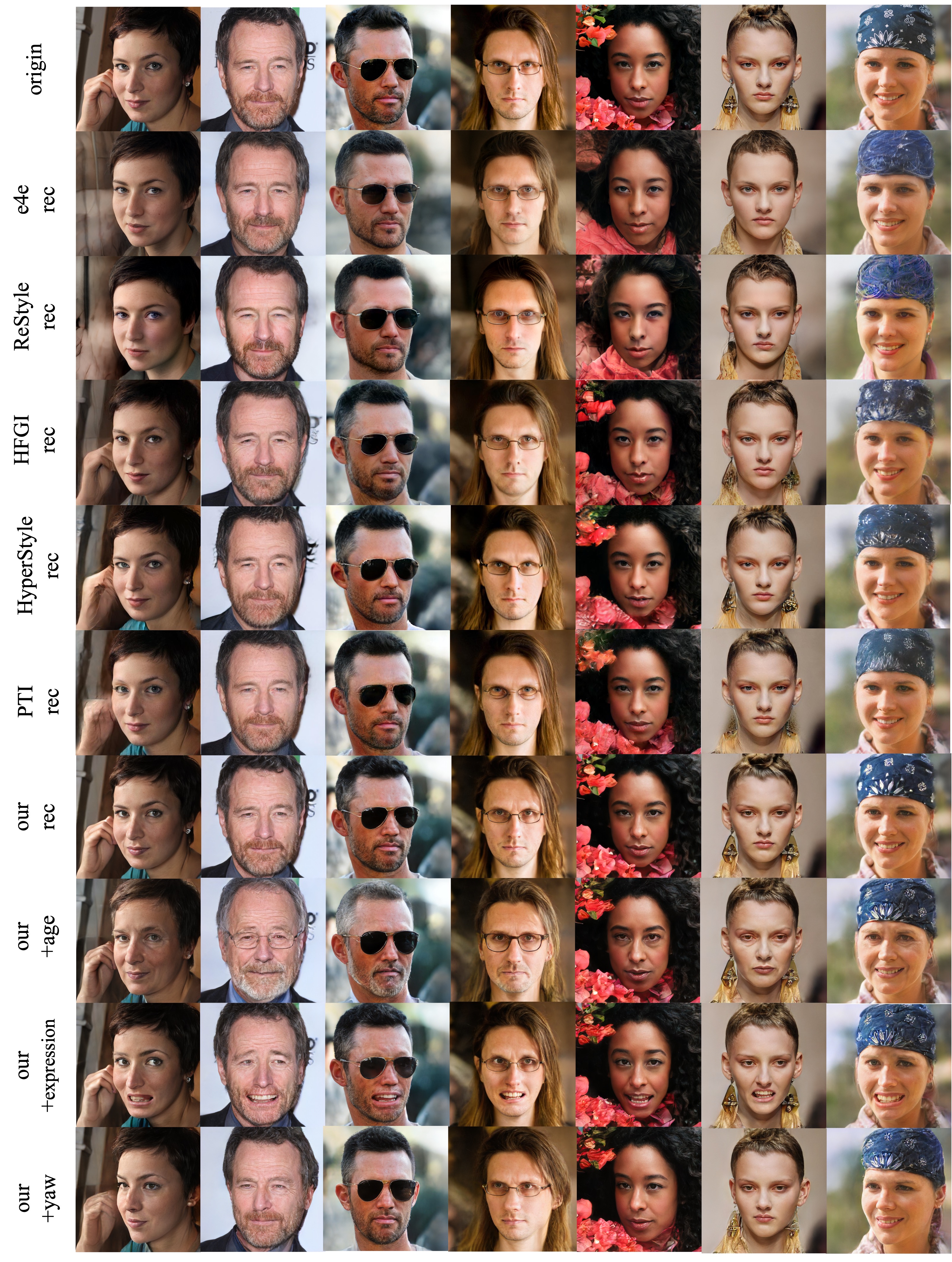}
	\centering
	\caption{More visual results on the human face dataset \cite{karras2019style}.}
	\label{fig:sup_face}
\end{figure*}
\clearpage

\begin{figure*}[p]
	\includegraphics[width=\linewidth]{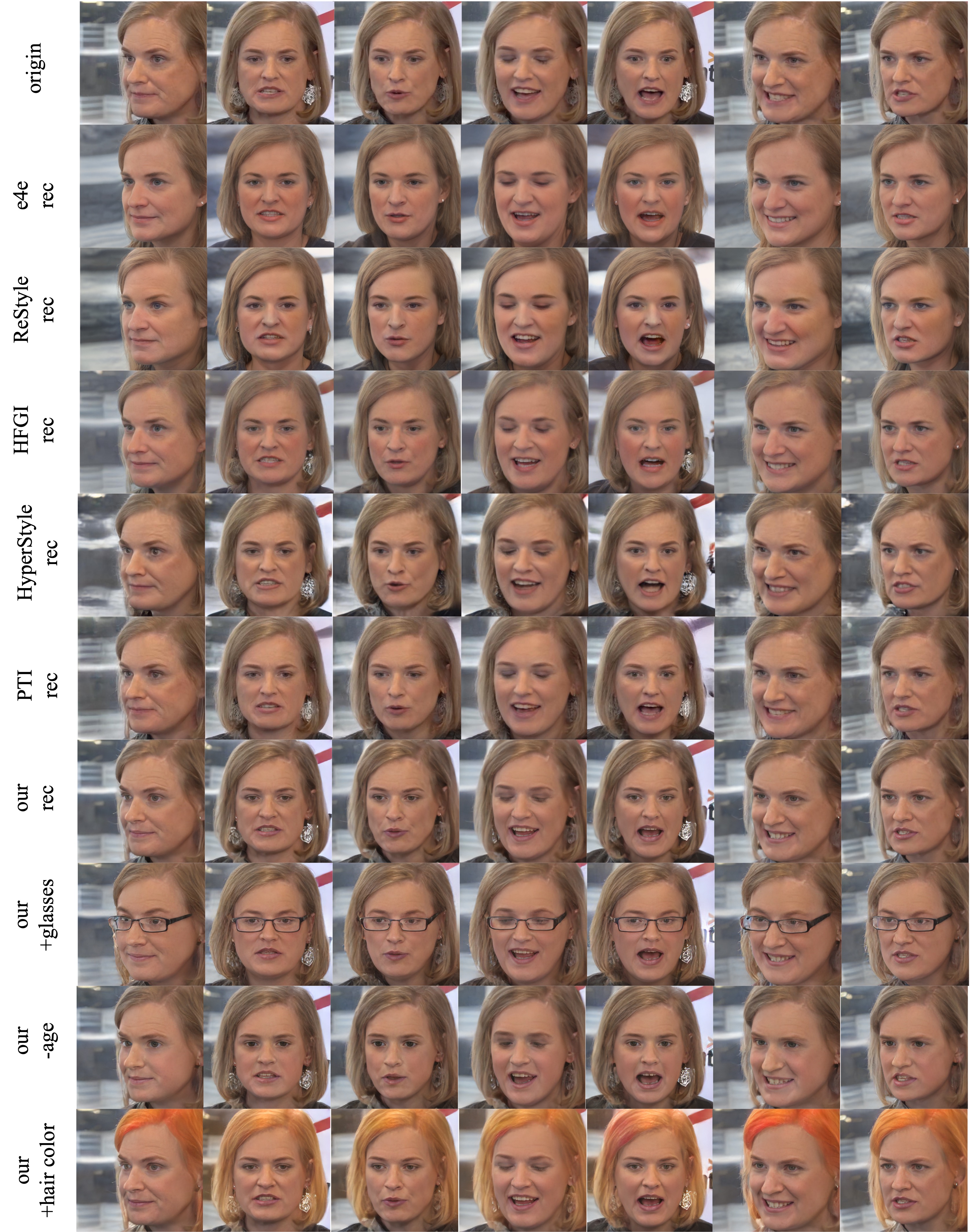}
	\centering
	\caption{Reconstruction and editing results on each frame of a real video \cite{wang2021one}.}
	\label{fig:sup_video}
\end{figure*}
\clearpage

\begin{figure*}[h]
	\includegraphics[width=\linewidth]{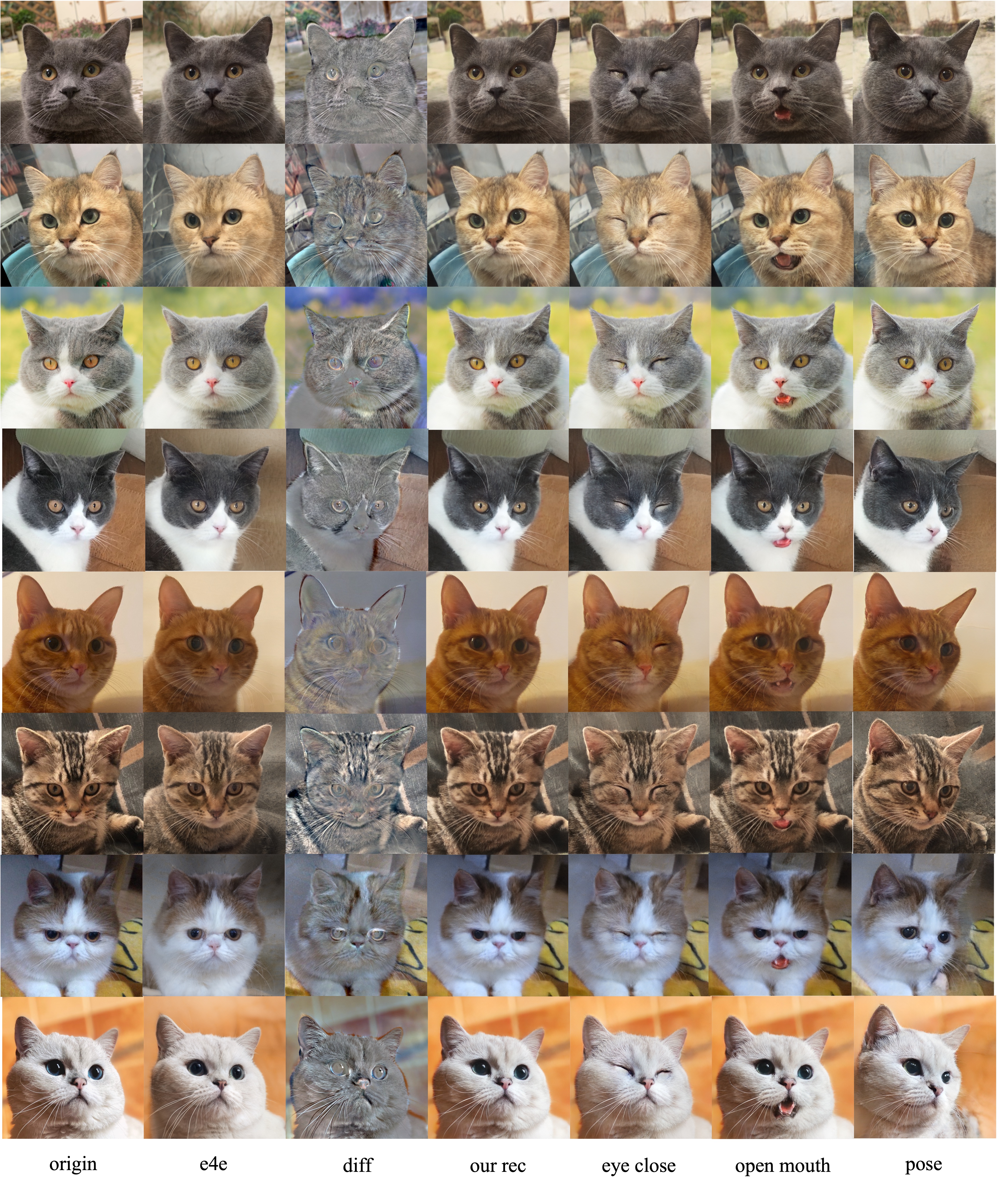}
	\centering
	\caption{Reconstruction and editing results on the cat images \cite{choi2020stargan}.}
	\label{fig:sup_cat}
\end{figure*}

\clearpage 

\begin{figure*}[h]
	\includegraphics[width=\linewidth]{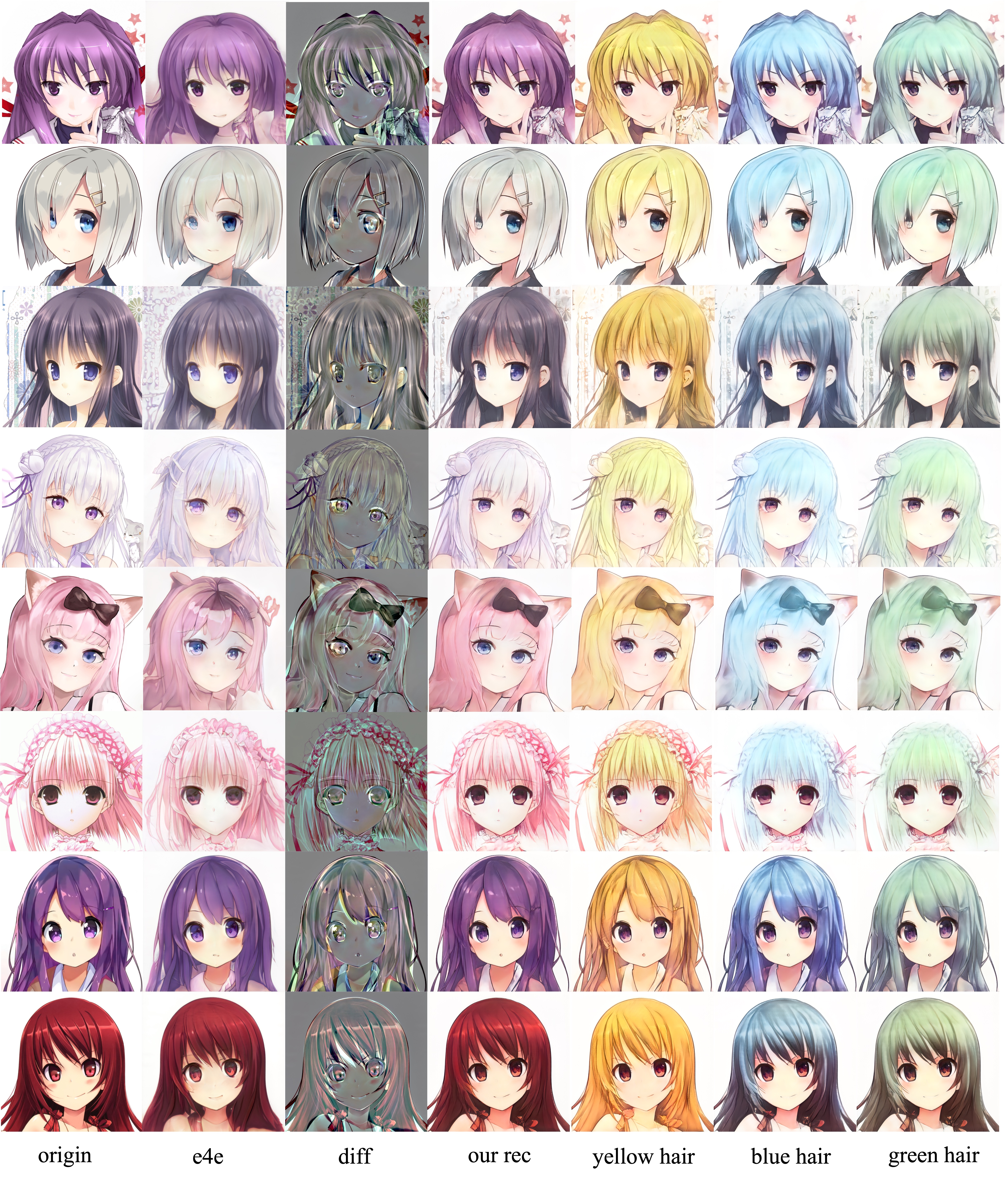}
	\centering
	\caption{Reconstruction and editing results on the anime images \cite{branwen2019danbooru2019}.}
	\label{fig:sup_anime}
\end{figure*}

\clearpage 

\begin{figure*}[htbp]
	\includegraphics[width=\linewidth]{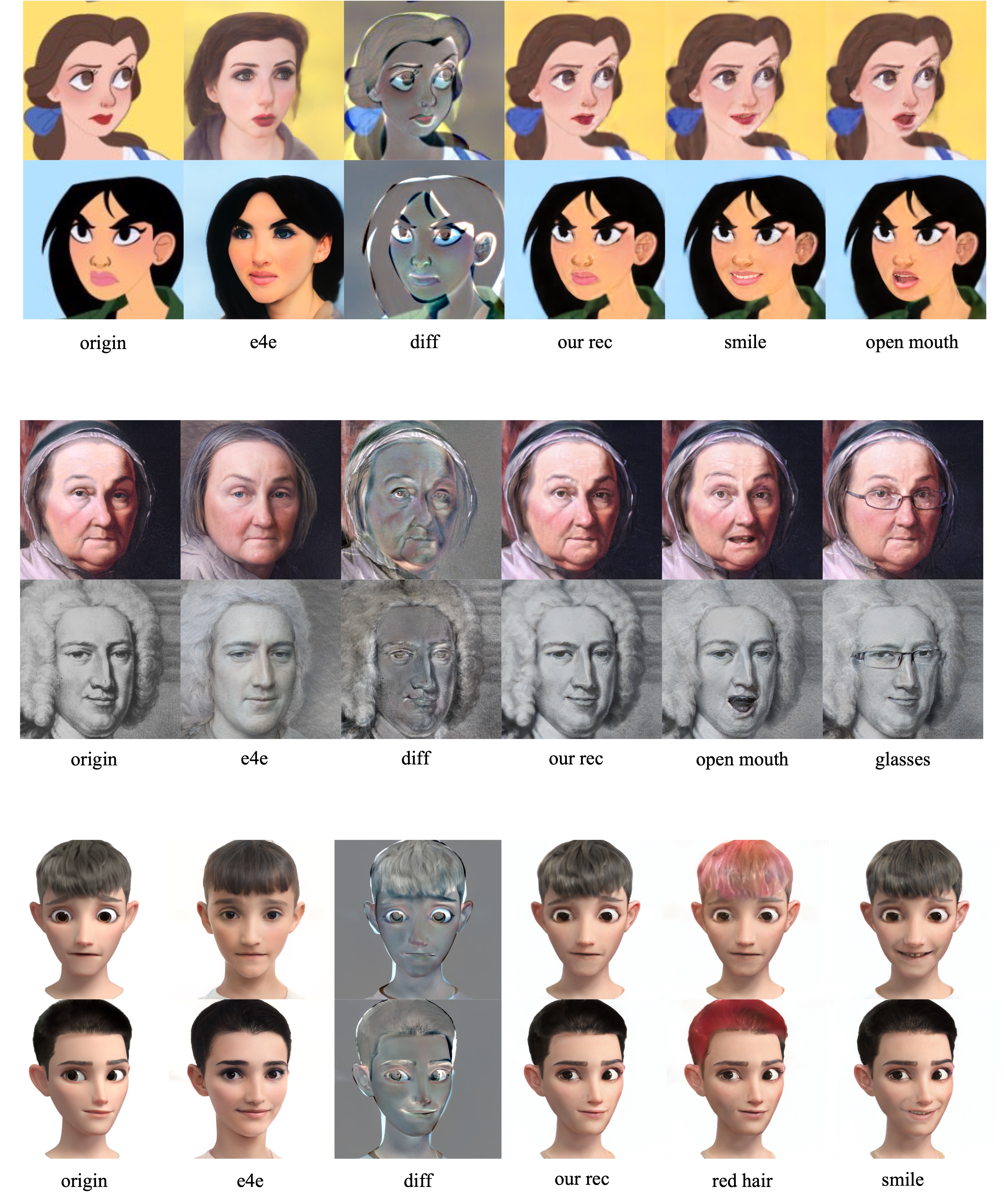}
	\centering
	\caption{Visualization of the out-of-domain rectifying results on the stylized portrait dataset \cite{pinkney2020resolution,karras2020training}. Note that the model is only trained on FFHQ dataset but can perform well on stylized portraits.}
	\label{fig:sup_domain}
\end{figure*}

\clearpage 

\begin{figure*}[t]
	\includegraphics[width=0.84\linewidth]{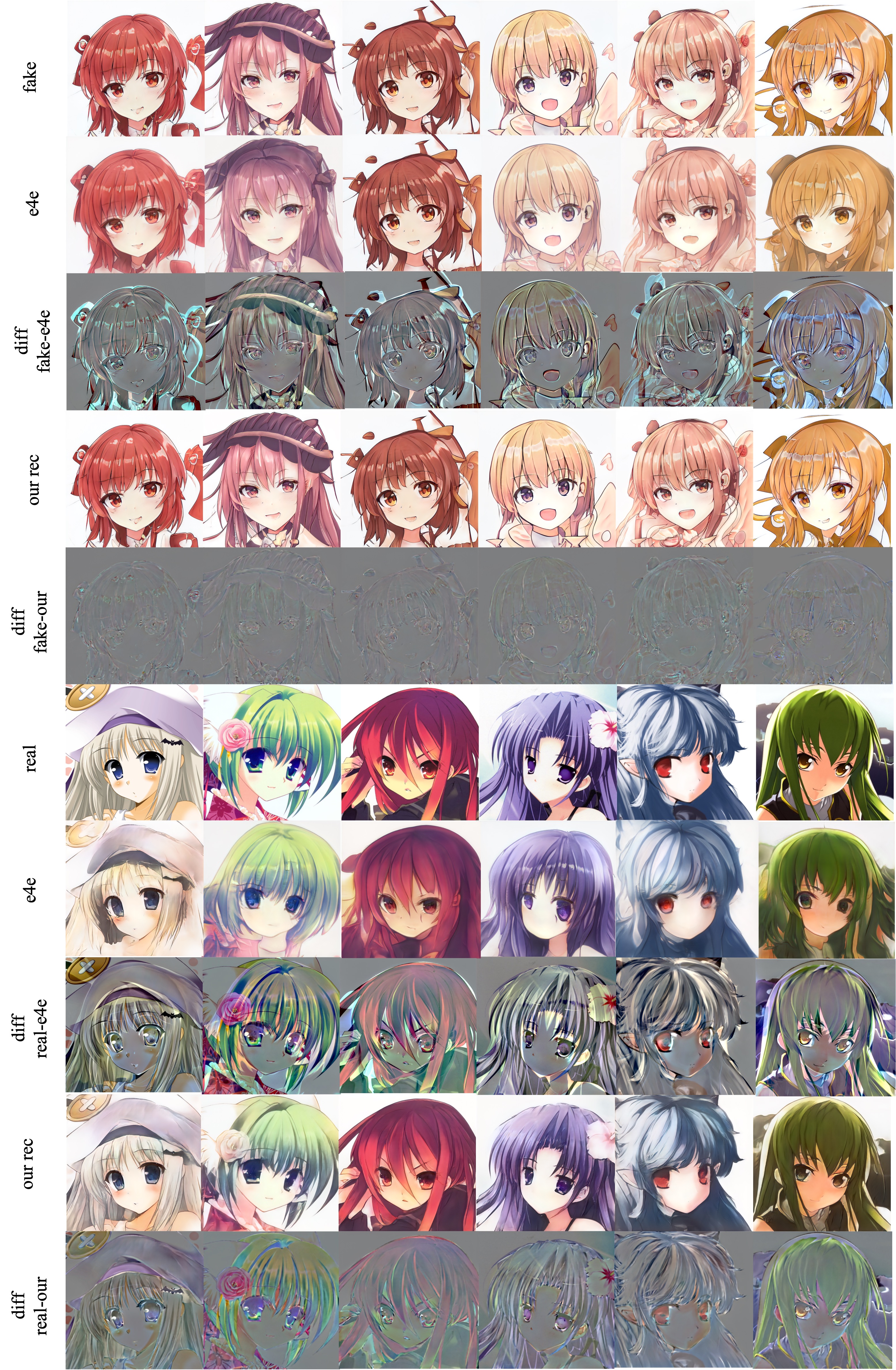}
	\centering
	\caption{Visualization of failure examples on the wild anime dataset \cite{branwen2019danbooru2019}, where the rectifying network fails to reconstruct the GAN inversion errors. Due to the distribution gap between the real anime data and the anime data synthesized by StyleGAN2 \cite{karras2020analyzing}, the chance of failure on real samples goes higher than that on the fake data.}
	\label{fig:sup_badcase}
\end{figure*}

\end{document}